\documentstyle[12pt,graphicx,natbib,ogonek]{article}

\begin{document}
\title{The Rank--Frequency Analysis for the Functional Style Corpora in the Ukrainian Language}
\author{Solomija N.~Buk$^*$, Andrij A.~Rovenchak$^{**}$\\
$^*$ Department for General Linguistics, Ivan Franko National University of Lviv,\\
1 Universytetska St., Lviv, UA-79000, Ukraine\\
$^{**}$ Department for Theoretical Physics, Ivan Franko National University of Lviv,\\
12 Drahomanov St., Lviv, UA-79005, Ukraine
}

\maketitle

\abstract{
We use the rank--frequency analysis for the estimation of Kernel Vocabulary size within
specific corpora of Ukrainian. The extrapolation of high-rank behaviour is utilized for
estimation of the total vocabulary size.

}

\section{Introduction}
The problem of rank--frequency analysis for texts is a very interesting one.
In application to the natural languages it gives a possibility to draw the information
which is necessary when compiling dictionaries, in particular, professionally-oriented
dictionaries, creating text-compressors, determining the basic vocabulary for studying
a language as foreign, etc.

In recent years, the development of computational techniques made it possible to study
large amounts of text. Such analysis usually involves the so-called Zipf's law \citep{Zipf49}
establishing the relation between the rank of a word and its frequency.
It was shown that the initially supposed linear behaviour on large samples of text
gets broken \citep{Nes87,Mon01,CanSol01}.

Several decades ago, statistical study of the Ukrainian language was held in the Potebnja
Institute of Linguistics in Kyiv \citep{Per67,Str74}. In these research works, however,
the results were just established, and no special analysis was made.
Unfortunately, such studies in Ukraine had been stalled for years, and only now they
are revived with application of modern techniques \citep{Dem01}.

In the paper we present results for different functional styles of Ukrainian language.
Such material is novel since the statistical analysis as well as corpus studies of Ukrainian
are now standing in the first stages only. While the volume of material involved in this
work is quite small comparing with e.~g., English, we hope that described techniques
together with preliminary results will be useful in future.

The paper is organized as follows. It the next section the description of
sources and text processing is given.
Section~3 contains the analysis of rank--frequency dependencies for different corpora
due to some specific features. The possible techniques for estimation of the vocabulary
size are adduced in Section~4. A brief discussion is presented in Section~5.

\section{Material Overview}

\subsection{Definition of Terms}
In this work we use the following terms:
\begin{itemize}
\item {\bf Corpus} --- body of collection of linguistic data, specially the one considered complete
      and representative, from a particular language or languages, in the form of recorded
      utterances or written text, which is available for theoretical or/and applied
      linguistic investigation \citep{Bur98}. In the present paper we consider
      {\bf text corpus} which must be distinguished from {\bf corpus of language (national corpus)}
      being a structured representative collection of texts from a given language.

\item {\bf Token} --- a word in any form (a sequence of letters between two spaces)
      in a text, e.~g., the sentence {\it I have not seen her yet} contains
      six tokens;

\item {\bf Corpus size} --- total number of tokens in the given corpus;

\item {\bf Vocabulary size} --- number of different words in the given corpus generated by the
      {\bf lemmatisation} process;

\item {\bf Lemmatisation} --- process of the reduction of word-forms to the initial (vocabulary)
      form, e.~g., verbs to the Infinitive, nouns to Nominative Singular, etc.

\item {\bf Vocabulary volume} --- estimated number of possible different words of the language
      (in the content of this work we mean it within specific functional style);
\end{itemize}

\subsection{Corpus Description}
In this work, we analyse a middle-sized corpus of Ukrainian language.
The size classification of corpora uses the Brown Standard Corpus of American English
\citep{Brown} as a reference point. Its parameters are as
follows: a)~one million words of running text; b)~500 text samples;
c)~2 thousand words per sample.
Corpora with less than one million words are considered as small,
corpora with 1--10 million words are middle-sized, and corpora containing
more than 10 million words are large.

Total corpus size alalysed in this work is about 1.7 million tokens.
It consists of five sub-corpora according to main five functional styles of speech (genres).

1.~The sub-corpus of {Belles-lettres Prose} contains 500 thousand tokens.
The frequency data were taken from \citep{Per81}. This frequency dictionary was compiled
on the basis of 25 creative works, with several text pieces extracted from different
places of one work. Although the time of the writings is restricted to 1945--1970, we suggest
that the changes in the first three thousand most frequent words are not significant.

2.~The sub-corpus of {\it Colloquial Style} contains about 300 thousand tokens.
It consists of 45 text pieces over approximately 6,000 tokens each.
Since big collections of `pure' Ukrainian colloquial speech do not exist, we used modern
dramas written within the last two decades~\citep{Buk03a}. The adequacy between these two
types of speech might be disputable but such a principle was used, e.~g., in \citep{JuiBro70}
and \citep{KurLew90}.

3.~The sub-corpus of {\it Scientific Style} was collected from 104 pieces each containing
about 3,000 tokens. Its total size slightly exceeds 300 thousand tokens.
The following scientific areas were represented in approximately equal
parts: biology, chemistry, psychology and pedagogics, physics, mathematics, technics,
geography and geology, history, linguistics~\citep{Buk03b}.

4.~{\it Official (business) Style} corpus was composed from texts of different kinds of
documents. These are: The Constitution of Ukraine, codices, Ukrainian and international laws,
international treaties, conventions, memoranda, declarations, speeches, economic documents,
contracts, all types of administrative documents, etc.
The size of the sub-corpus is about 300 thousand tokens.

5.~{\it Journalistic Style} frequency statistics was taken from \citep{UkrPub}.
The correspondent corpus build on basis texts from several all-Ukrainian newspapers
issued in 1994. These newspapers are addressed to both city-dwellers and villagers,
and to people of different age. The size of the sub-corpus is also about 300 thousand tokens.

\subsection{Text processing}
At the first stage, several types of items were removed from texts. These are: numbers,
word containing numbers, punctuation signs (see comment on dashes below),
and words written in a non-Ukrainian script.
Then, texts were processed manually for homonyms. This is a very important stage as some
of these words appear with high frequency. As an example, we propose some homonym pairs
(note, that stress is usually omitted in Ukrainian\footnote{Hereafter for the sake of convenience we use
transliteration for representing Ukrainian words according to the table given in Appendix.}):
{\it br\'aty} (`to take', verb in the Infinitive) and {\it brat\'y} (`brothers', noun
in Plural, Nominative); {\it m\'aty} (`to have', verb in the Infinitive) and
{\it m\'aty} (`mother', noun in Singular, Nominative); {\it ni\v{z}} (`than', particle)
and {\it ni\v{z}} (`knife', noun in Singular, Nominative); {\it \v{s}\v{c}o} being a particle,
a conjunction (`which'), and a pronoun (`what'); etc.

\section{Frequency Analysis}

\subsection{Low ranks}

The behaviour at low ranges is significantly influenced by some
specific features of Ukrainian language. Several very frequent
words have different forms due to the principle of so called
euphony. Namely, the word
{\it i} (`and') may appear also in the
forms {\it j} and {\it ta}. The word {\it v} (`in') may have also
forms {\it u} and {\it vvi} or {\it uvi} (the last two are rare).

The verb `to be', very frequent in different language corpora, in Ukrainian can be
replaced by a dash (---) or omitted at all, and therefore, it appears a bit less
frequently when comparing with its rank in other languages, especially in spoken language.
Note, however, that the inverse statement is incorrect. i.~e., not every dash represents
this verb.

In the table below we present first five most frequent words from different corpora.
English statistics is based on the British National Corpus \citep{BNC}.
German language statistics was kindly granted by Sabine Schulte from
the University of Stuttgart. Croatian corpus data is taken from \citep{HNC}, and Polish
is from \citep{PWN}. Ukrainian statistics is collected by the authors.

\bigskip
\noindent
\begin{center}
\begin{tabular}{c|ll|ll|ll|ll|ll}
\hline
\hline
Rank&\multicolumn{2}{c|}{English}&\multicolumn{2}{c|}{German}
    &\multicolumn{2}{c|}{Croatian}&\multicolumn{2}{c|}{Polish}
    &\multicolumn{2}{c}{Ukrainian}\\
\hline
1& the &0.0619& die  &0.0702& i  &0.0314& w       &0.0317& i  &0.0371\\
2& be  &0.0424& sein &0.0289& u  &0.0276& i       &0.0282& v  &0.0303\\
3& of  &0.0309& in   &0.0274& je &0.0264& si\k{e} &0.0192& na &0.0173\\
4& and &0.0268& der  &0.0245& se &0.0156& na      &0.0167& z  &0.0166\\
5& a   &0.0219& ein  &0.0234& da &0.0130& z       &0.0159& ne &0.0157\\
\hline
\hline
\end{tabular}
\end{center}


\bigskip
This table demonstrates that our data are consistent with other Slavic languages.

\subsection{Kernel Vocabulary}
Zipf formulated the relation between the frequency of the word $f$ and its rank $r$,
basing on the `principle of least effort' which he considered as one of the most
important features of human behaviour, on the analogy of Poincar\'e's principle
of least action in physics. A slightly modified, in comparison with its original form,
this dependence reads:
\begin{equation}\label{Zipf}
f_r=A/r^z,
\end{equation}
where $A$ and $z$ are parameters, the exponent $z$ slightly deviates from unity.
(Originally Zipf put the value $z=1$). Further, we refer this relation as Zipf's law.


We have analysed rank--frequency dependencies for our corpora in the following way.
Since the Zipf's law (\ref{Zipf}) after taking the logarithm from both sides is linearised,
it is common to express the rank--frequency relations in a log--log plot
(see figures below).

\bigskip
\centerline{\includegraphics[angle=-90,width=70mm,clip]{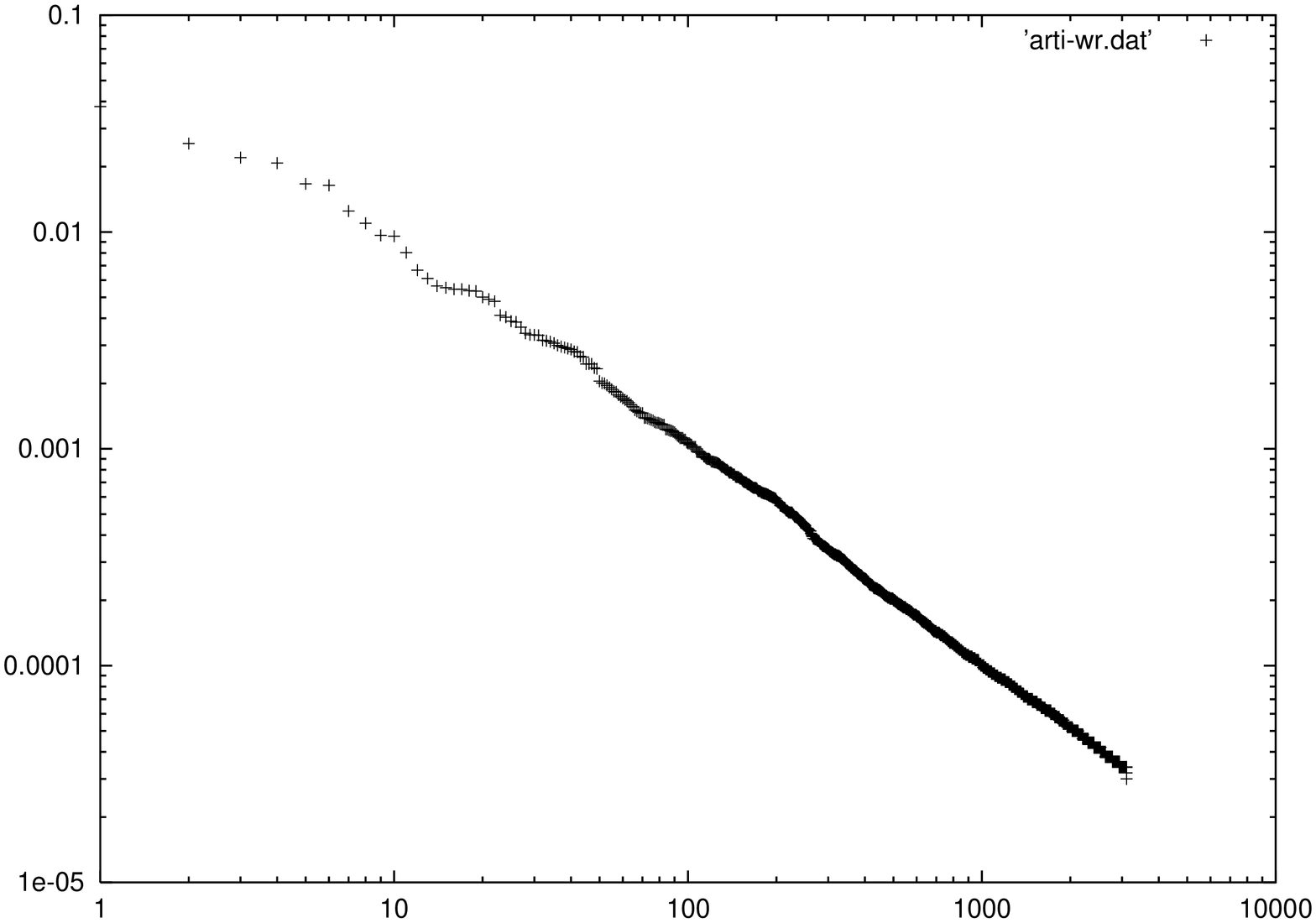}\
\includegraphics[angle=-90,width=70mm,clip]{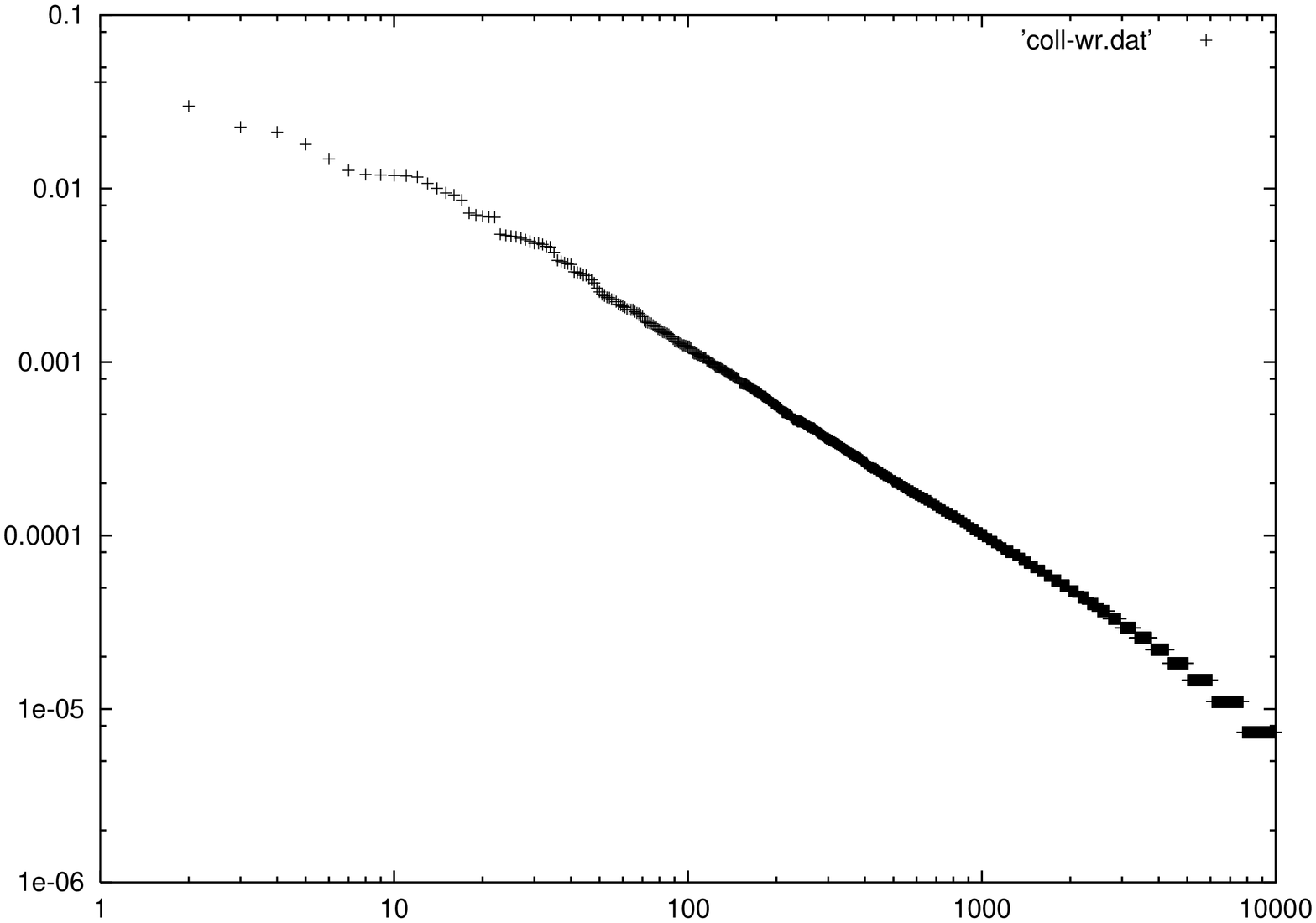}}
\centerline{Belles-letres \hfil Colloquial}
\centerline{\includegraphics[angle=-90,width=70mm,clip]{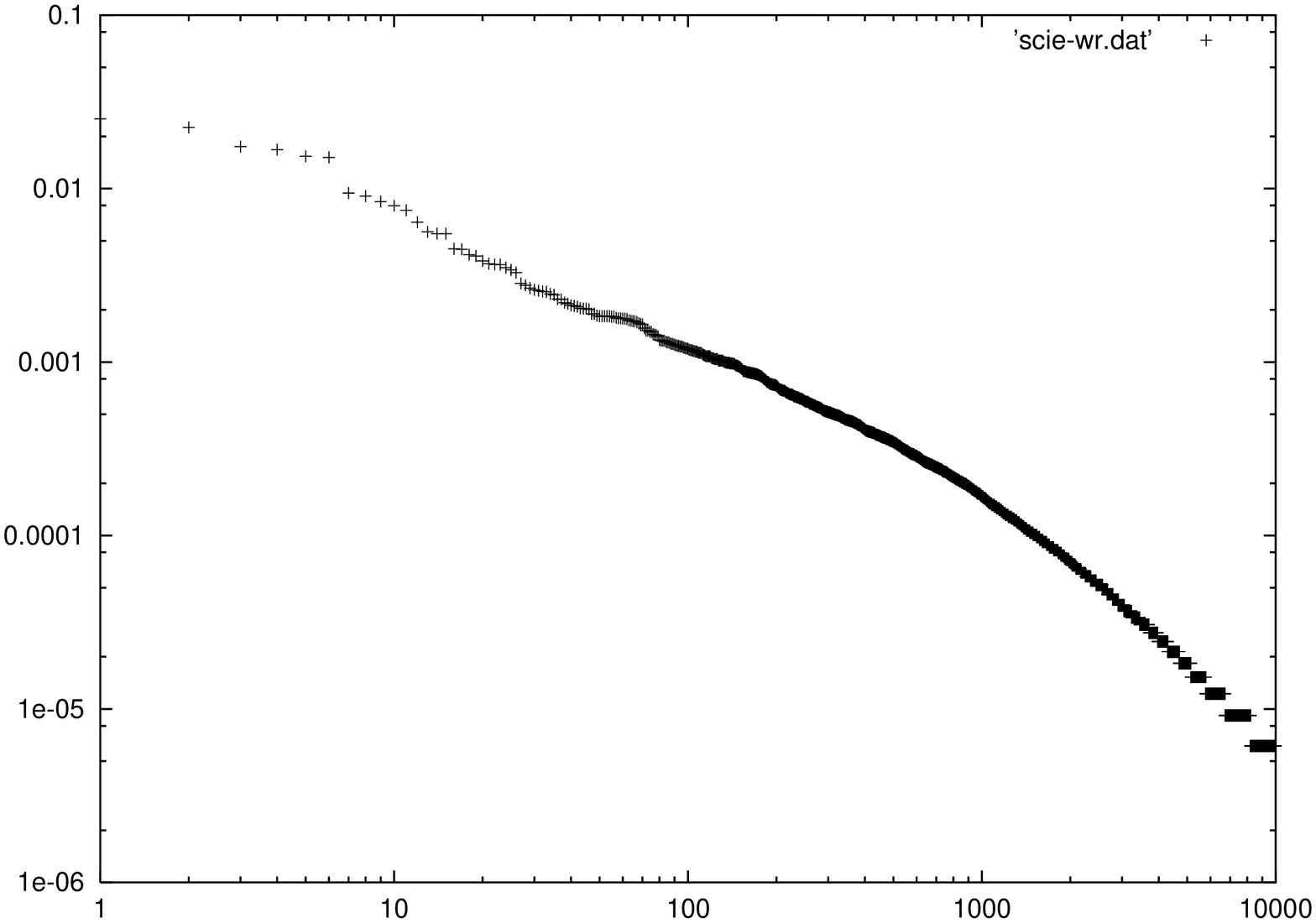}\
\includegraphics[angle=-90,width=70mm,clip]{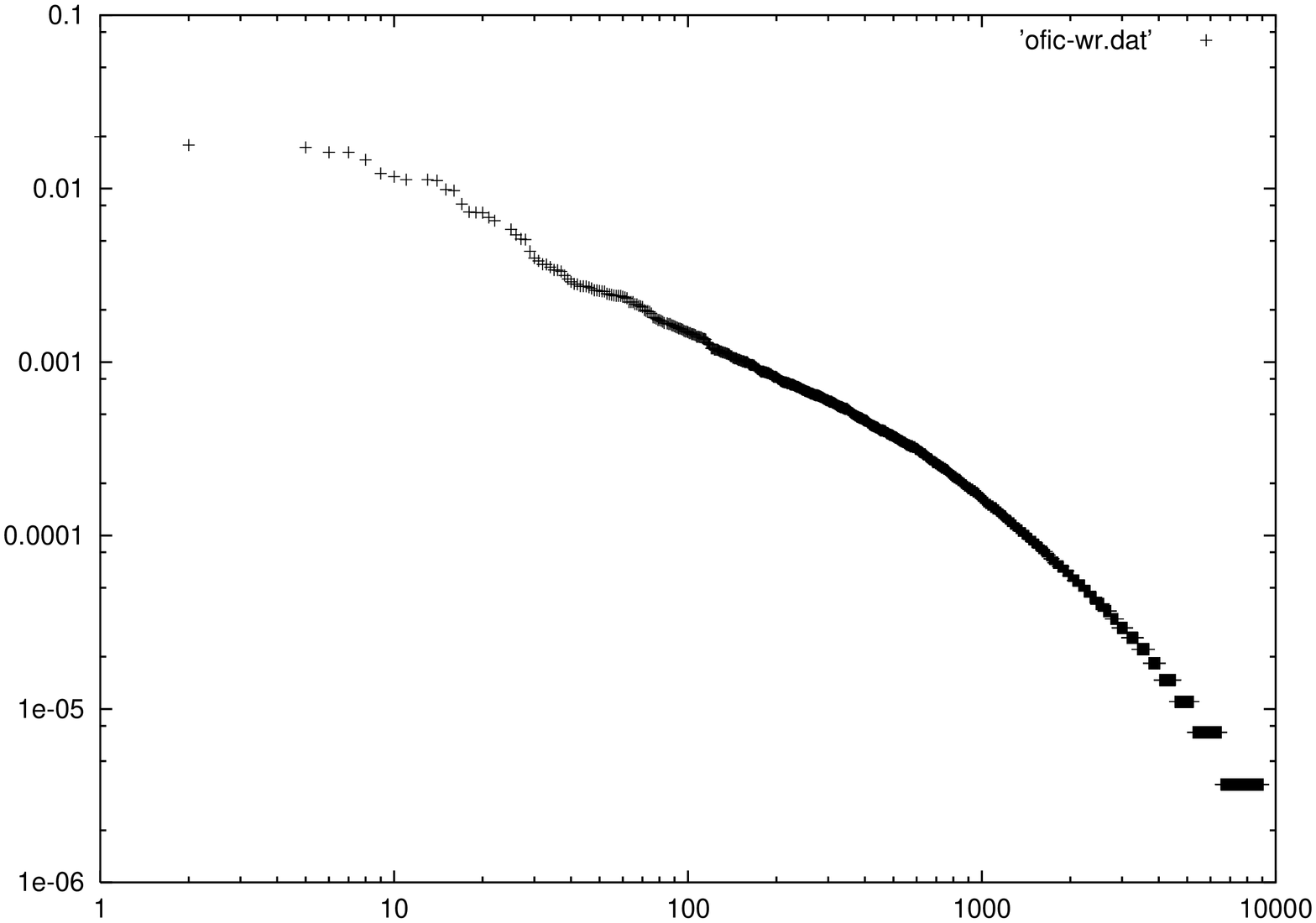}}
\centerline{Scientific \hfil Official}
\centerline{\includegraphics[angle=-90,width=70mm,clip]{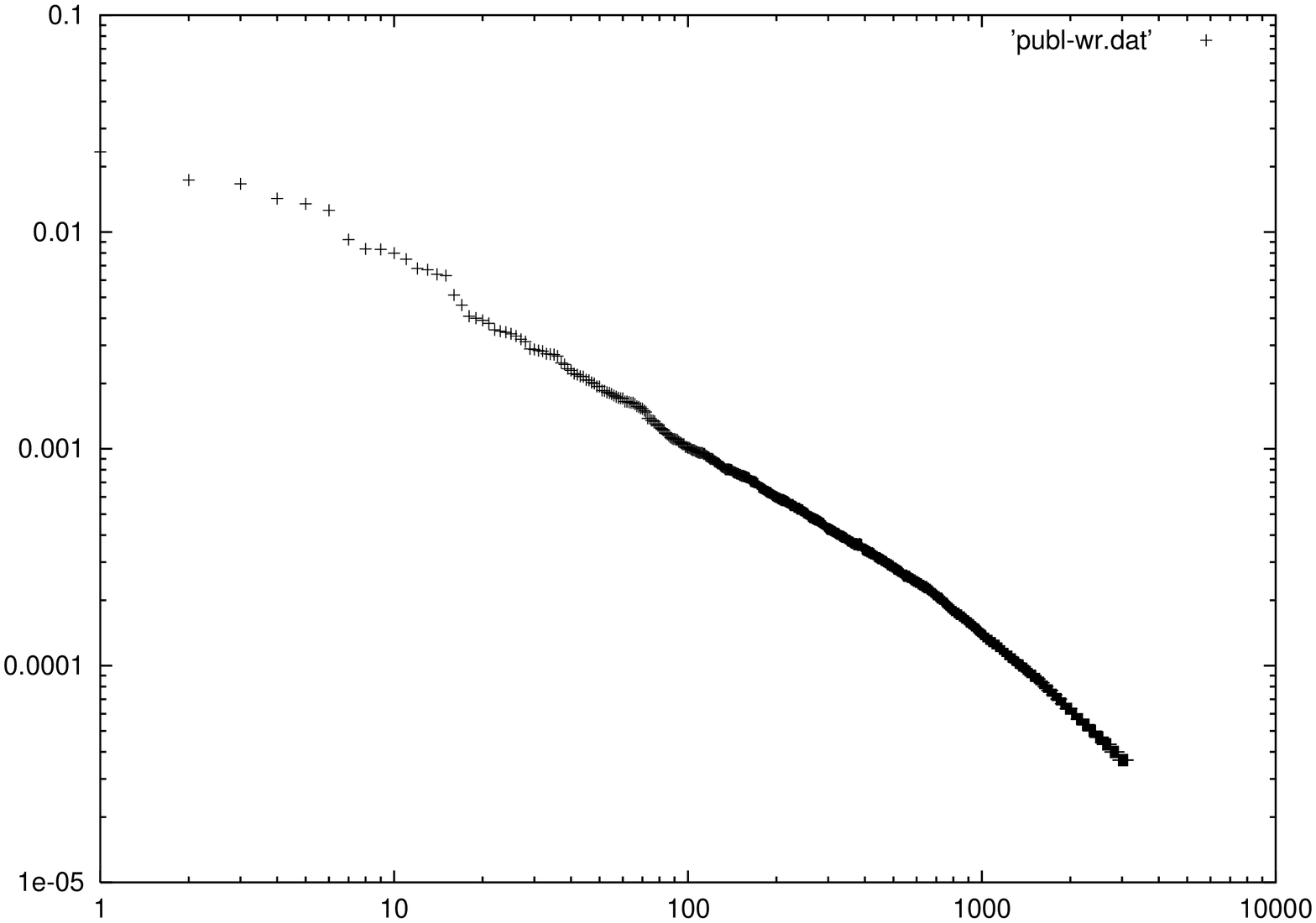}\
\includegraphics[angle=-90,width=70mm,clip]{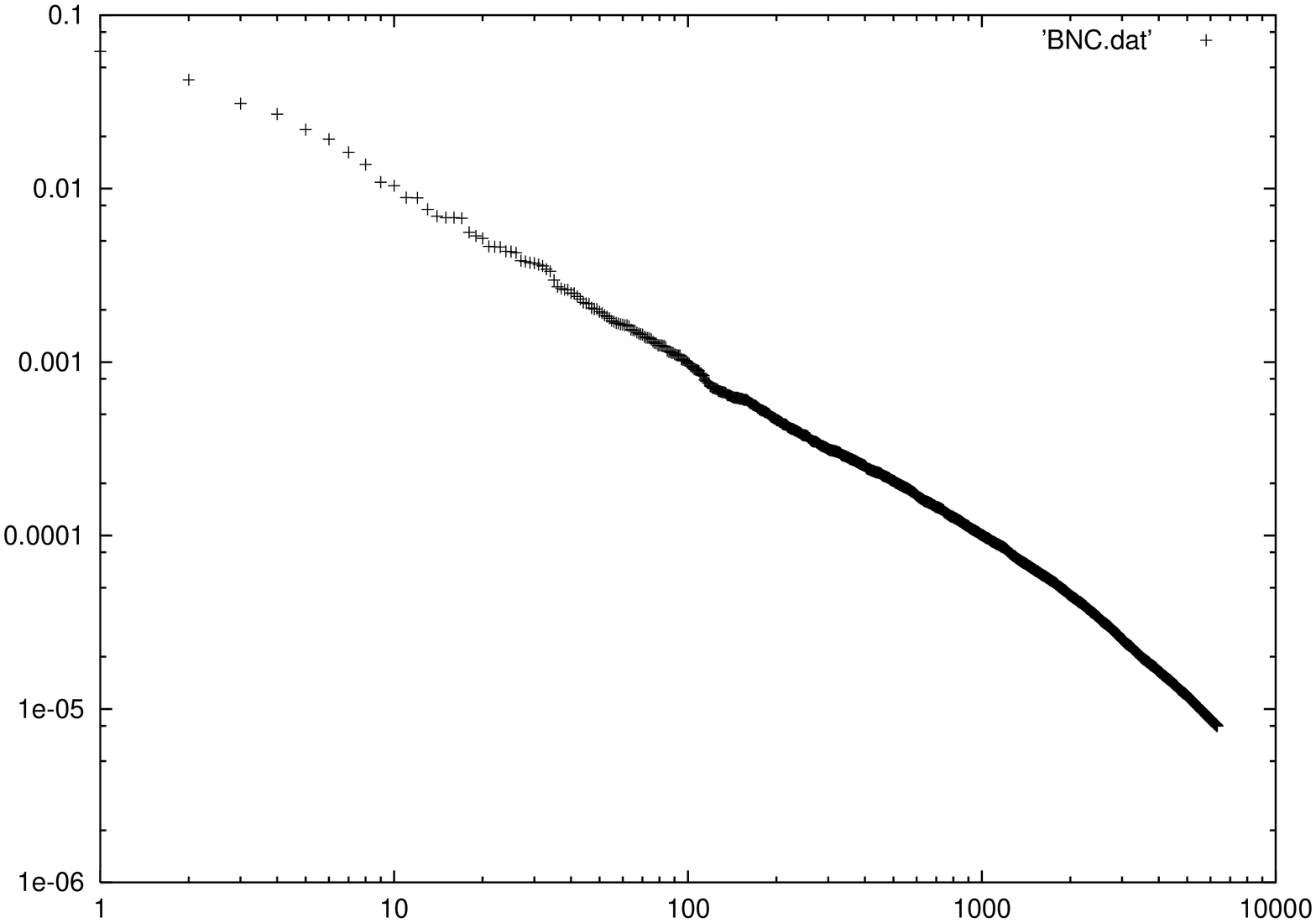}}
\centerline{Journalistic \hfil BNC}
\bigskip

The idea of the selection of Kernel Vocabulary is based on the assumption
that on the `rank--frequency' curve the deviation from linear (Zipf's) behaviour corresponds
to the transition to a different type of vocabulary \citep{Mon01}.
The author made the analysis using the British National Corpus \citep{BNC}.
Although the size of our corpus is far from the size of British National Corpus
but anyway, as we show further, such scales already allow for conclusions
on some statistical features of the text under consideration.

One can easily notice a slight change in the curve slope when moving to higher-rank region.
In order to find the place where this change occurs a detailed analysis is required.
We have divided the ranks into domains of 200: from 1 to 200, from 101 to 300,
from 201 to 400 and so on.
Then, for each domain the best-fit parameters to Zipf's law (\ref{Zipf}) were calculated.

After making the detailed numerical analysis of data for each sub-corpus
we noticed the following specific features:
\begin{itemize}
\item in the official, journalistic and scientific sub-corpora at some rank $r_{\rm max}$
      the value of $z$
      changes significantly, which corresponds to the transition to a different part
      of the vocabulary. The values are $r_{\rm max}\simeq800$ for the official sub-corpus,
      $r_{\rm max}\simeq1000$ for the scientific one,
      $r_{\rm max}\simeq 1600$ for journalistic sub-corpus.
\item in the colloquial corpus the deviation from (\ref{Zipf}) is less significant,
      the Zipf's law with $z=1.09$ describes the whole domain of ranges quite well. However,
      the numerical analysis allows for stating the value of $r_{\rm max}$ close to that
      of the journalistic sub-corpus.
\end{itemize}

In order to give a better understanding for the behaviour of the Zipf's exponent
we propose a visual interpretation in Fig.~\ref{ExpVisual} below.

\begin{figure}[h]\label{ExpVisual}
\centerline{\includegraphics[angle=-90,width=100mm,clip]{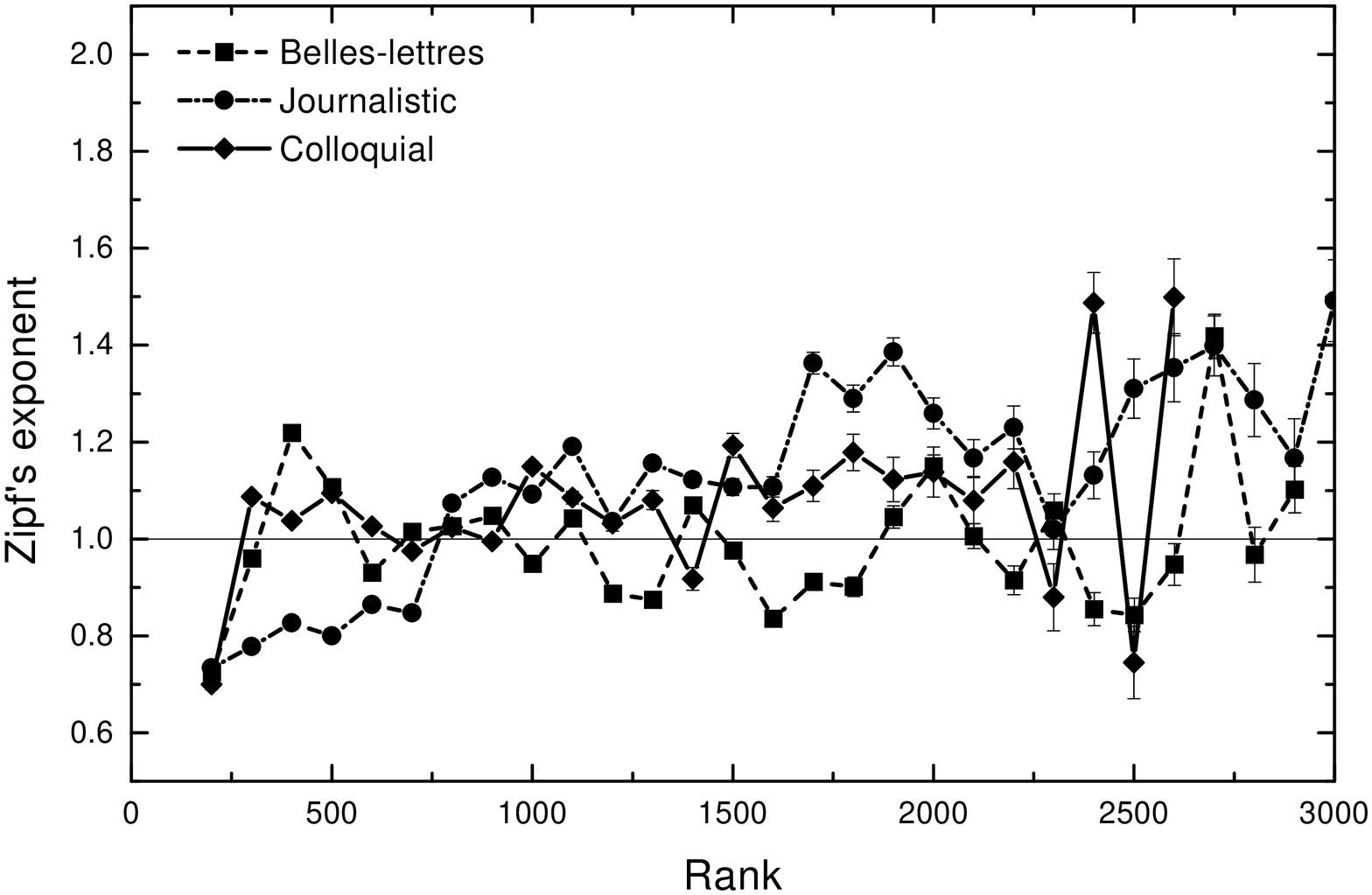}}
\centerline{\includegraphics[angle=-90,width=100mm,clip]{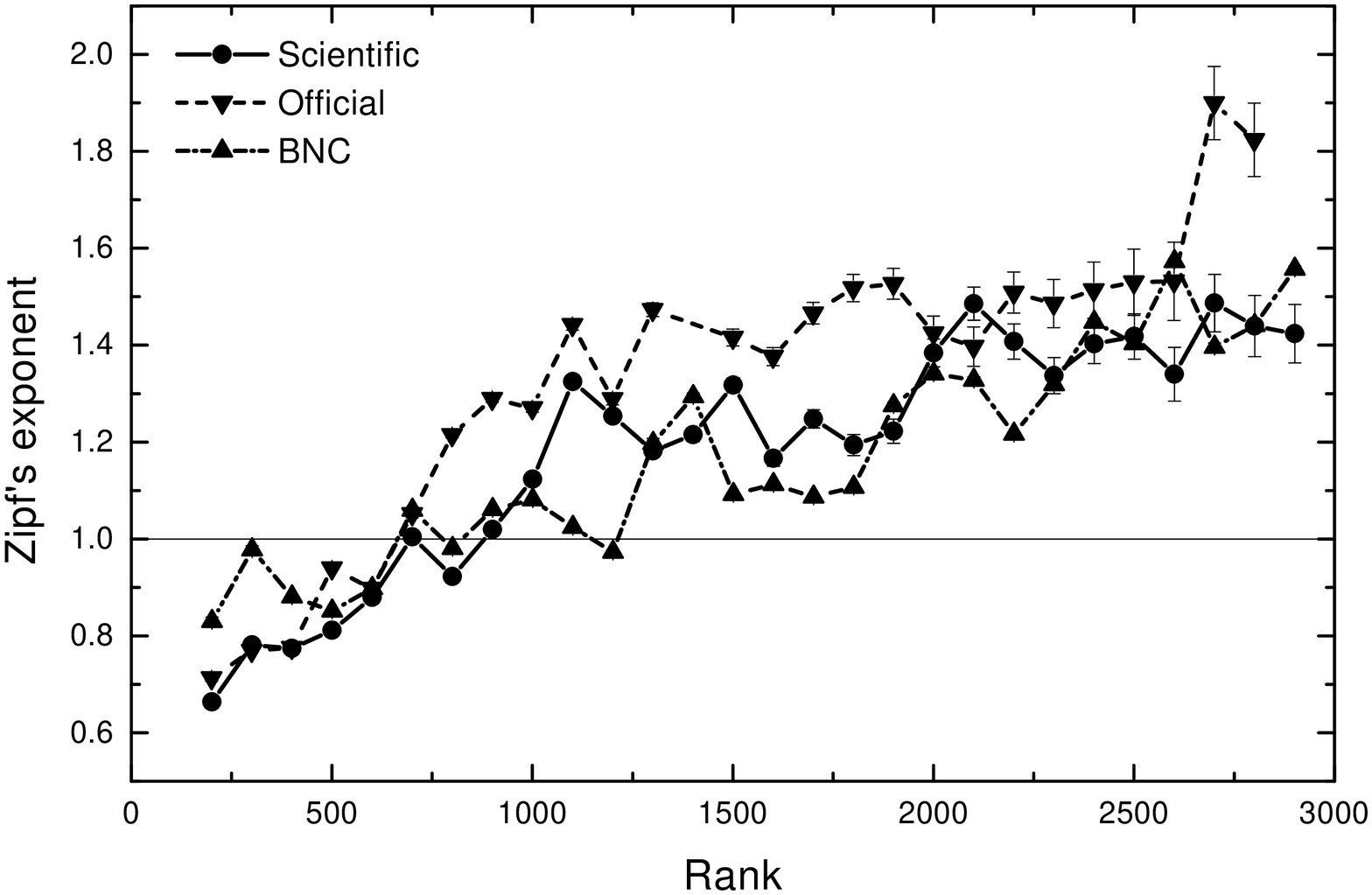}}
\caption{Zipf's exponent behaviour showing the transition between different types of
vocabulary.}
\end{figure}

\subsection{Entropy Comparison}
It is interesting to analyse the frequency dependencies due to the entropy $S$:
\begin{equation}
S_N=-\sum_{r=1}^{N} f_r\,\ln f_r,
\end{equation}
where $N$ is a big number. By putting $N=3000$ for each sub-corpus we obtained
the following values: BP 2.192; CS 2.356; PS 2.368; SS 2.602; OS 2.750.
While the smallest value of entropy for the belles-lettres sub-corpus looks a bit unexpected,
we propose the following interpretation of the rest data. In physics, the entropy
is the measure of disorder in a system. As we know from our experience, official texts are
usually hardly-readable, therefore, they need more effort to be understood. In scientific
texts a similar statement is a bit less applicable when taking into account the fact of
reading the text by addressees --- specialists in the respective field. From this point of
view, the journalistic texts must be quite close to the everyday speech --- and we see it
from the numbers.

\section{Vocabulary size estimation}
Suppose one has the whole language corpus, and its vocabulary size is $\cal R$.
This means that $\cal R$ is the maximal possible rank, so frequency $f$ of the
next-ranked word $f_{{\cal R}+1}$ will be zero. If one accepts Zipf's dependence
(\ref{Zipf}) to be valid, such situation never appears. Let us therefore accept a bit
modified function \citep{Lua94}:

\begin{equation}
f_r^t=-A+B r^t,
\end{equation}
where the exponent $t$ is a small positive number.
In this case, the value of $\cal R$ is defined as follows: ${\cal R}=(A/B)^{1/t}$.
Typical values of $t$ are of order 0.1. Thus, the estimation of the vocabulary size for the
specific functional genre gives the values 200 to 700 hundred different words. A more precise
estimation will be made after larger corpus is analysed.

\section{Discussion}

We analysed the rank--frequency relations for the middle-sized corpus
of the Ukrainian language. The data for the first most ranked words are consistent with other
Slavic languages. The presented results allows for the establishing of the
Kernel vocabulary and vocabulary size estimation. The entropy was calculated for different
functional genres.
We hope that our data will be useful when compiling the National corpus
of the Ukrainian language. A more precise results will be available after larger corpus
is considered.

\section*{Appendix}

\bigskip
\begin{figure}[h]
\centerline{\includegraphics[width=120mm,clip]{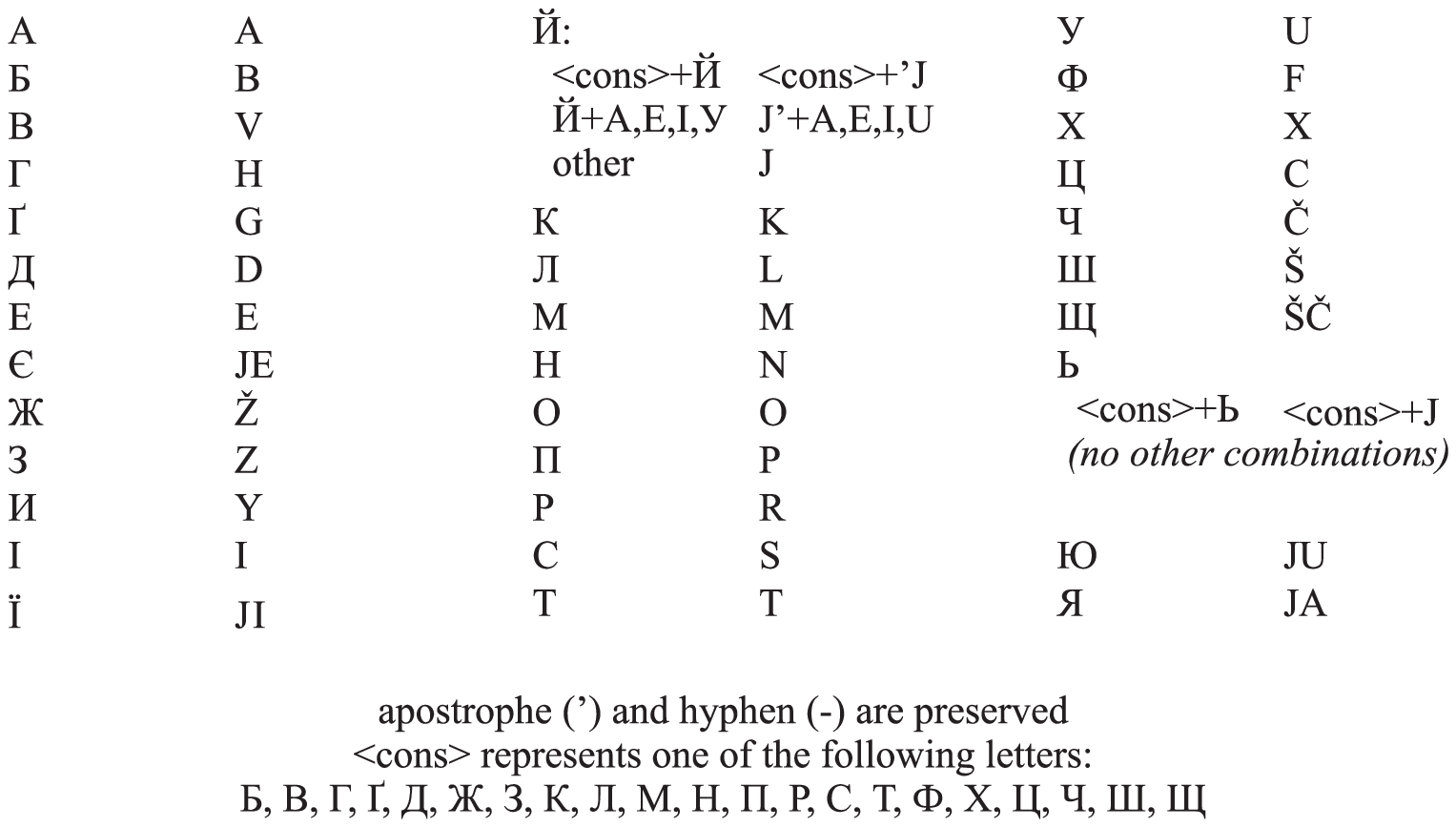}}
\caption{Ukrainian Transliteration Table.\protect\\
This transliteration scheme is free of ambiguity and allows for making bi-directional
transliterations. While in some places it seems to be a bit complicated, in the practical
applications difficult letter combinations appear very rarely. In addition, it
is concordant with some Slavic written systems based on Latin script.}
\label{TranslitTable}
\end{figure}

\end{document}